\documentclass[letterpaper]{article} % DO NOT CHANGE THIS
\usepackage{aaai2026}  % DO NOT CHANGE THIS
\usepackage{times}  % DO NOT CHANGE THIS
\usepackage{helvet}  % DO NOT CHANGE THIS
\usepackage{courier}  % DO NOT CHANGE THIS
\usepackage[hyphens]{url}  % DO NOT CHANGE THIS
\usepackage{graphicx} % DO NOT CHANGE THIS
\usepackage{natbib}  % DO NOT CHANGE THIS AND DO NOT ADD ANY OPTIONS TO IT
\usepackage{caption} % DO NOT CHANGE THIS AND DO NOT ADD ANY OPTIONS TO IT
\usepackage{algorithm}
\usepackage{algorithmic}

\usepackage{booktabs}
\usepackage{amssymb}
\usepackage{xcolor}
\usepackage{amsmath}

\usepackage{newfloat}
\usepackage{listings}
\DeclareCaptionStyle{ruled}{labelfont=normalfont,labelsep=colon,strut=off} % DO NOT CHANGE THIS
\floatstyle{ruled}
\newfloat{listing}{tb}{lst}{}
\floatname{listing}{Listing}
\title{CapeNext: Rethinking and Refining Dynamic Support Information for Category-Agnostic Pose Estimation}
\author{
    Yu Zhu\textsuperscript{\rm 1,2},
    Dan Zeng\textsuperscript{\rm 1}\thanks{
      Corresponding author: zengd8@mail.sysu.edu.cn
    },
    Shuiwang Li\textsuperscript{\rm 3},
    Qijun Zhao\textsuperscript{\rm 4},
    Qiaomu Shen\textsuperscript{\rm 5},
    Bo Tang\textsuperscript{\rm 2}
}
\affiliations{
    \textsuperscript{\rm 1}School of Artificial Intelligence, Sun Yat-sen University, Zhuhai 510275, China\\
    
    \textsuperscript{\rm 2}Department of Computer Science and Engineering, Southern University of Science and Technology, Shenzhen 518055, China\\
    
    \textsuperscript{\rm 3}College of Computer Science and Engineering, Guilin University of Technology, Guilin 541004, China\\

    \textsuperscript{\rm 4}College of Computer Science, Sichuan University, Chengdu 610065, China\\
    
    \textsuperscript{\rm 5}Aerospace and Informatics Domain, Beijing Institute of Technology, Zhuhai 519088, China\\
    yuzhu83@foxmail.com, zengd8@mail.sysu.edu.cn, lishuiwang0721@163.com, qjzhao@scu.edu.cn, joyshen06@gmail.com, tangb3@sustech.edu.cn
}

\usepackage{bibentry}
\begin{document}
 
\maketitle

\begin{abstract}
Recent research in Category-Agnostic Pose Estimation (CAPE) has adopted fixed textual keypoint description as semantic prior for two-stage pose matching frameworks. While this paradigm enhances robustness and flexibility by disentangling the dependency of support images, our critical analysis reveals two inherent limitations of static joint embedding: (1) polysemy-induced cross-category ambiguity during the matching process(e.g., the concept "leg" exhibiting divergent visual manifestations across humans and furniture), and (2) insufficient discriminability for fine-grained intra-category variations (e.g., posture and fur discrepancies between a sleeping white cat and a standing black cat). 
To overcome these challenges, we propose a new framework that innovatively integrates hierarchical cross-modal interaction with dual-stream feature refinement, enhancing the joint embedding with both class-level and instance-specific cues from textual description and specific images. Experiments on the MP-100 dataset demonstrate that, regardless of the network backbone, CapeNext consistently outperforms state-of-the-art CAPE methods by a large margin.
\begin{links}
\link{Code}{https://github.com/yzrs/CapeNext}
\end{links}
\end{abstract}   
\section{Introduction}
\label{sec:intro}
\begin{figure}[h]
  \centering
    \includegraphics[width=\linewidth]{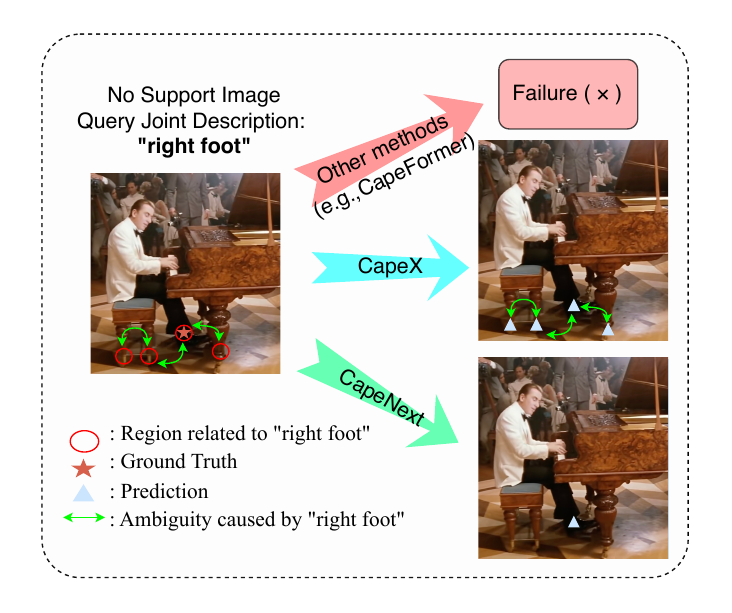}
    \caption{Conventional CAPE methods can't work without support image. With only textual keypoint descriptions, CapeX will be affected by the textual ambiguity to match the wrong candidate points, while our method works better.}
    \label{fig:1900}
\end{figure}
Category-Agnostic Pose Estimation (CAPE), as a fundamental task in computer vision, has demonstrated significant values in application ranging from action recognition to human-computer interaction for arbitrary category  \cite{minderer2022simple,zhang2023simple,zareian2021open}. 
POMNet \cite{xu2022pose} reformulates category-agnostic pose estimation as a keypoint matching problem, where the model learns to transfer keypoint knowledge from annotated support images to query images of unseen categories. 
Subsequent researches \cite{shi2023matching,hirschorn2023pose,chen2024meta,ren2024dynamic} train CAPE models in a two-stage framework, generating similarity-aware position proposals and then refining them through a transformer decoder. However, the model performance in this paradigm is affected to a greater extent by the quality of the support image, and even when there are invisible keypoints in the support image, it fails to provide effective keypoint knowledge for the query image. To address this issue, CapeX \cite{rusanovsky2025capex} replaces unstable keypoint knowledge from support images with stable text feature of keypoint descriptions encoded by LLM. 

However, compared to support images, relying solely on textual keypoint descriptions can lead to inherent ambiguity in complex scenarios. For example, in Figure \ref{fig:1900}, when searching for ``right foot", there are multiple candidates that match the query, which cannot be effectively distinguished using text alone.
Let's take a step back and consider how humans approach this task. When searching for the ``right foot", we subconsciously focus on the pianist's shoes, making it unlikely for us to mistake the chair's leg for a foot. That's because we instinctively associate ``right foot" with concepts like ``human", ``leg", and ``shoes" using them as additional contextual cues to guide our final decision while naturally disregarding the chair's foot. 
However, obtaining rich relevant cues for CAPE is both cumbersome and impractical for model training. 
Beyond additional associative information, humans naturally perceive the overall context of an image, even when not specifically searching for \mbox{keypoints}. 

Inspired by this, we propose leveraging the perceptual understanding of the query image as contextual support and utilizing easily accessible category information as associative information. 
The former introduces the visual features of the query image and reduces the difficulty of matching between the support text and the query image in the case of large intra-category instance variance. It adds the detail information from query images to keypoint features to help the keypoints dynamically adapt to different query images. The latter provides us with class information that can help filter the noise information of other non-target objects in the query image, enhance the feature expression of the subject and diminish polysemy-induced cross-category ambiguity. It is worth noting that we are the first to incorporate the query image itself as additional ``support" information to construct keypoint embeddings. 

However, effectively integrating the query image, class description, and keypoint description to improve CAPE is challenging, primarily due to the modality differences between class descriptions and query image features. Direct feature fusion can also introduce noise, adversely affecting model performance. This is because the query image contains relevant target information and extraneous noise. Moreover, the detailed features in class descriptions may conflict with specific attributes of the query image, further complicating effective fusion.

To this end, we propose \textbf{CapeNext} to elevate CAPE's performance to the \textbf{Next} level by the strengths of different modalities while mitigating potential feature conflicts. Specially, CapeNext contains two innovative modules: Hierarchical Cross-Modal Interaction~(HCMI) and Dual-Stream Feature Refinement~(DSFR). 
These two modules align the feature from both the specific image and the general class representation, leveraging the images' strengths in capturing fine-grained, instance-level details and text's ability to encode broad, coarse-grained semantics, refining the keypoint
feature from different aspects.
CapeNext introduces multimodal input and innovative feature refinement modules to dynamically adopt the joint embedding of the fixed keypoint descriptions to be closer to the specific query image. 

To sum up, this paper makes the following contributions:
\begin{itemize}
    \item We rethink the support information in CAPE and propose CapeNext, which is the first to incorporate query information as ``support" to facilitate the dynamic feature learning of keypoints.
    \item We introduce two novel modules, Hierarchical Cross-Modal Interation (HCMI) and Dual-Stream Feature Refinement (DSFR), which effectively leverage the class-level and instance-specific cues to produce constant text-modal joint embedding that resolves polysemy-induced cross-category ambiguity and enhances discriminability for fine-grained intra-category variations.
    \item Our experiments on the MP-100 dataset demonstrate that, regardless of the network backbone, CapeNext consistently outperforms state-of-the-art CAPE methods.
\end{itemize}

\section{Related Work}
\label{sec:relatedWork}

% \subsection{Category-Agnostic Pose Estimation}
% Different from conventional pose estimation methods \cite{cao2019openpose,yang2023explicit,sun2019deep,xiao2018simple,ding20222r,li2021tokenpose,andriluka2018posetrack}, Category-Agnostic Pose Estimation~(CAPE) eliminates category restrictions during testing, enabling pose prediction for arbitrary target classes. This generalization capability aligns closely with real-world application demands, where models must adapt to diverse, unseen categories without requiring category-specific training data.
\textbf{Category-Agnostic Pose Estimation (CAPE)} removes category restrictions during testing, enabling pose prediction for arbitrary targets, different from conventional pose estimation methods \cite{cao2019openpose,yang2023explicit,sun2019deep,xiao2018simple,ding20222r,li2021tokenpose,andriluka2018posetrack}. Its generalization aligns with real-world needs, where models must adapt to diverse unseen categories without category-specific training data.

POMNet \cite{xu2022pose}, the first category-agnostic method, frames pose estimation as embedding-space semantic matching between support keypoint and query image features but suffers from unreliable results due to visually similar irrelevant regions (e.g., mirror keypoints) and insufficient similarity under large pose/texture/style differences.
CapeFormer \cite{shi2023matching} pioneers a two-stage transformer framework: generating similarity-aware initial keypoint proposals via improved matching, then refining them by aggregating context through Transformer Decoder.
X-Pose \cite{yang2024x} leverages CLIP \cite{radford2021learning} vision-language info via support visual prompts and structured text prompts ("An [style] photo of [object]/[part]/[keypoint]"), generating semantic embeddings to boost generalization to unseen categories via flexible prompt use.
SDPNet \cite{ren2024dynamic} models joint dependencies via skeleton graphs and GCN (with self-attention for adjacency matrices) but has suboptimal efficiency due to separate GCN training.
GraphCape \cite{hirschorn2023pose} replaces transformer feed-forward layers with GCN for end-to-end training (reducing overhead). CapeX \cite{rusanovsky2025capex} uses CLIP text prompts to match query features, cutting support image dependency and improving accuracy.

% These existing methods all follow a similar learning paradigm that enriches the representation of semantic keypoints in the support image by using text prompts of skeleton graphs to enhance CAPE accuracy. However, they ignore the fact that the query image also contains rich semantic information that can be used in return to refine the support information. Our method makes more full use of the information in the query image and dynamically adjusts the features of semantic keypoints provided in the support samples accordingly.
These existing methods share a paradigm of enriching support image keypoint representations via skeleton graph text prompts to boost CAPE accuracy, but overlook the query image's rich semantic information that can refine support data. Our method fully leverages query image information and dynamically adjusts support sample keypoint features accordingly.

\section{Our Method: CapeNext}
\subsection{Preliminaries}
\label{sec:introCAPE}
\begin{figure*}[h]
  \centering
    \includegraphics[width=\linewidth]{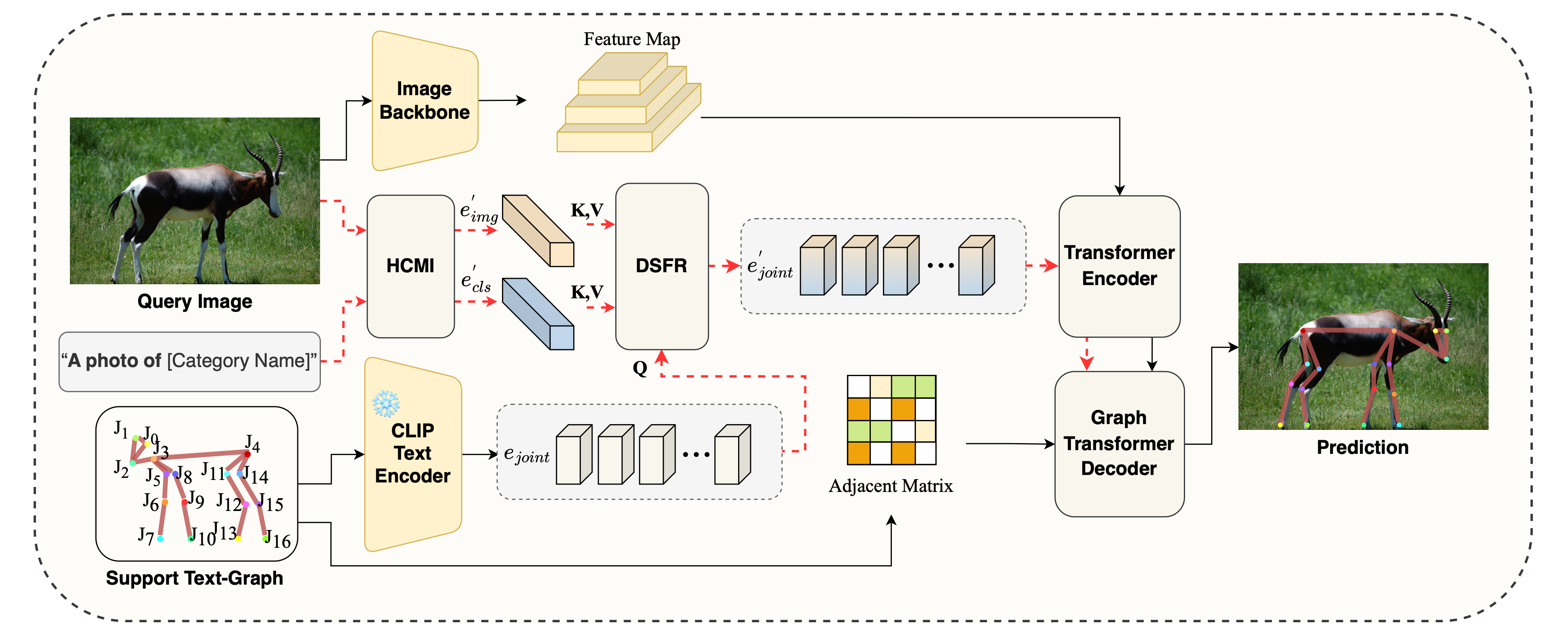}
    \caption{Framework Overview of CapeNext. CapeNext employs a dual-stream architecture to process multimodal inputs.}
    \label{fig:mainFig}
\end{figure*}

% 先介绍CAPE任务
% 之后最好再加上一个CAPE的公式
Category-Agnostic Pose Estimation aims to estimate desired keypoints on the query image given a support image annotated with keypoints for arbitrary category. Formally, under the 1-shot setting (one support image), given a support set $S=\{I_s,K_s\}$ where $I_s \in \mathbb{R}^{H\times W \times 3}$ denotes the support image and $K_s = \{k^j_s\}^N_{j=1}$ represents its corresponding $N$ keypoint coordinates, CAPE predicts the target keypoint set $\hat{K_q} = \{\hat{k}^j_q\}^N_{j=1}$ from a query image $I_q \in \mathbb{R}^{H\times W \times 3}$. The core idea is to leverage explicitly defined keypoint features from support samples to guide the model's search for similar feature patterns in query images, thereby disentangling category-specific prior knowledge dependencies. This process could be formulated as follows:
\begin{equation}
    \label{eq:CAPE-function}
    K_q = M_{\theta}(I_q,\{I_s,K_s\}),
\end{equation}
where $K_q$ indicates the predicted keypoints and $M_{\theta}$ is the CAPE model used here.

% 再介绍CapeX的优势与不足
% 考虑移动到intro中
% However, this approach leads to the CAPE's prediction accuracy being significantly influenced by support sample quality and noise contamination. As evidenced in the experimental results of prior methods, 5-shot configurations generally outperform 1-shot scenarios. 

CapeX propose a novel way for CAPE by replacing support keypoint feature with LLM-encoded text embedding, guiding the model to search for visual feature that are similar to text embedding in query images. Formally, it can be described as follows:
\begin{equation}
    \label{eq:CapeX-function}
    K_q = M_{\theta}(I_q,\{K^{text}_{s}\}),
\end{equation}
where $K^{text}_{s}\!=\!\! {\{K^{text_{j}}_{s}\}}^{N}_{j=1}$ indicates the support joint embedding obtained from LLM encodings.

This paradigm relies solely on textual keypoint descriptions rather than specific support images, eliminating dependency on support image quality. It is useful for unseen categories where annotated support samples are typically unavailable, demonstrating superior practicality in real-world applications. Simultaneously, benefiting from the semantic extensibility of textual descriptions, this paradigm based on textual keypoint representations also demonstrates great flexibility. Our method integrates this framework, inheriting the advantages of representational stability and input flexibility inherent in text-driven approaches.

\subsection{Motivation}
Previous methods benefit from the flexibility of using joint textual descriptions as support, but they inherit a fundamental drawback: text cannot capture fine-grained visual details as rich as images. For example, the textual representations ``tail" or even ``tail of a cat" can't accurately match a black cat's tail or a white cat's tail at the same time without a more detailed visual representations.

In the CAPE domain, its reliance on unconstrained textual keypoint definitions also introduces two fundamental limitations:(1) \textbf{Polysemy-induced Cross-category Ambiguity}: identical textual descriptions (e.g., “left leg”) may refer to anatomically distinct parts across categories (e.g., human legs vs. chair legs), causing the model to conflate heterogeneous geometric structures due to their semantic overlap in the text embedding space.(2) \textbf{Intra-category Variance}: even within a single category, instances with diverse appearances (e.g., a standing black cat’s ``right front paw" vs. a curled white cat’s ``right front paw") are forced to share the same text embedding, suppressing the model’s ability to recognize the pose-dependent visual variations.

% \begin{figure}[h]
%   \centering
%     \includegraphics[width=\linewidth]{figure/intro_d.png}
%     \caption{An example of large intra-category instance variance. ``right front paw" of the cat is marked by the red star.}
%     \label{fig:catDemo}
% \end{figure}

As illustrated in Figure \ref{fig:mainFig}, our framework CapeNext consists of three parts: (1) multimodal feature extraction and refinement, (2) feature matching and graph-based structural reasoning, and (3) keypoint prediction. Our key contribution lies in the design and refinement of multimodal feature which enhance support joint feature with both class-level and instance-specific cues. Transformer encoder and graph transformer decoder share the same architectures as CapeX.

\subsection{Method Overview}
Given a query image, we first extract its multi-scale visual features using a network backbone. These features capture both local part details and global contextual patterns. To overcome polysemy-induced cross-category ambiguity and intra-category instance variation issues in CAPE, we derive an enhanced joint embedding through our multimodal input: textual class descriptions, which are easy to obtain, along with the query image itself as complementary information for keypoint and textual class descriptions. 

The effectiveness of multimodal input comes from two aspects. 
First, textual class descriptions can mitigate polysemy-induced cross-category ambiguity. By incorporating textual class descriptions, the model resolves semantic ambiguities arising from identical textual representations across different categories. For example, when multiple objects in an image share identical keypoint descriptions but differ semantically, textual class descriptions can act as discriminative filters. These descriptions suppress interference from the regions irrelevent to the targets while constraining the semantic positioning space for valid keypoints.
Second, visual feature integration from query images addresses the issue of intra-category instance variance. Consider the query images of black cat and white cat: Neither original joint embeddings nor textual class description encode the uncertain visual attributes such as the fur color. When the query image shifts between black cat and white cat, the fixed keypoint representations struggle to adapt to the changed visual feature. But the integration of query image embedding can compensate for the joint embedding inaccuracies caused by significant intra-category variations because the target feature is just in the query image itself. It provides complementary instance-aware details, which are missing in textual class descriptions, and thus effectively balancing semantic precision and visual fidelity.

\begin{figure*}[h]
  \centering
    \includegraphics[width=0.9\linewidth]{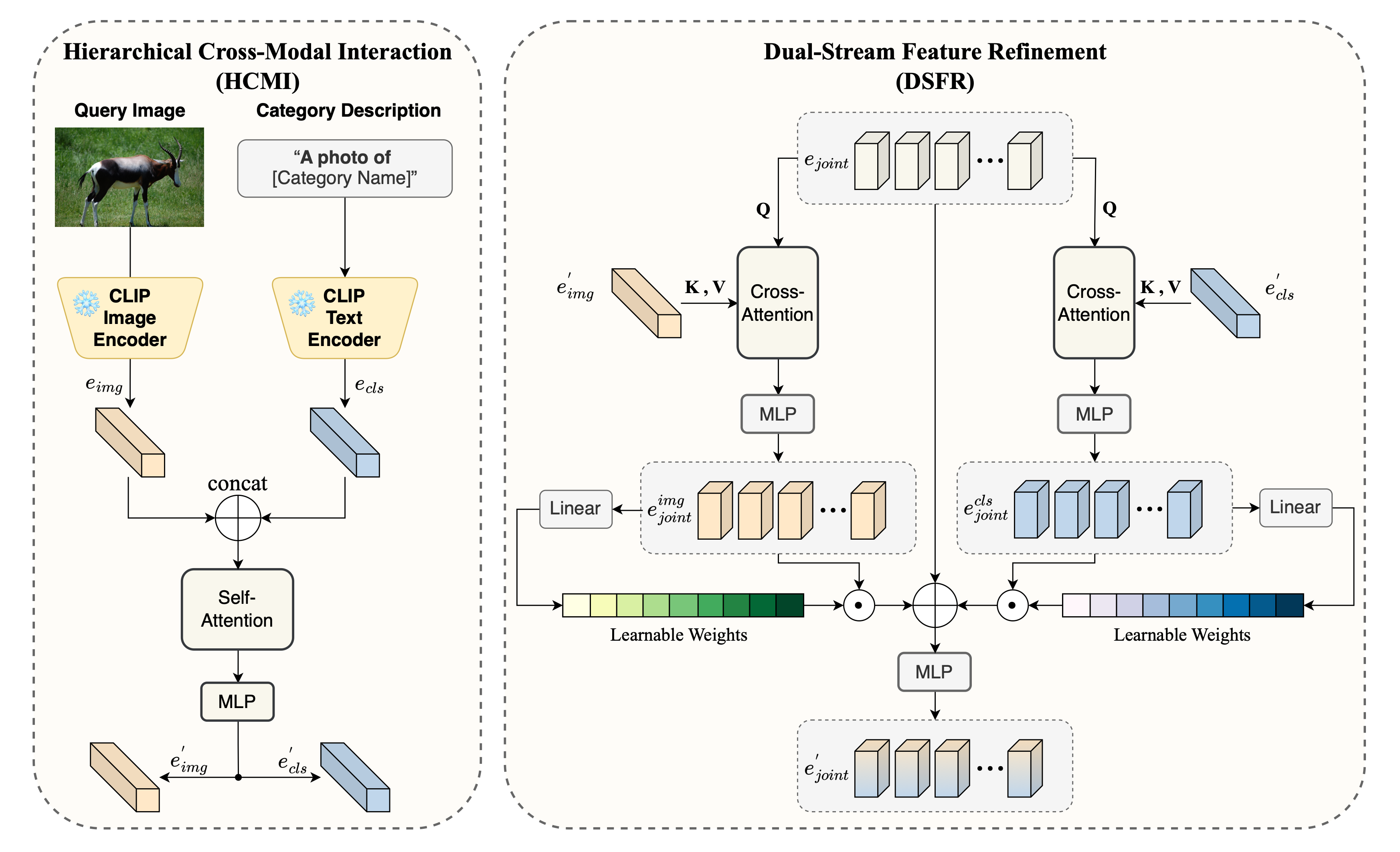}
  \caption{Left: HCMI mediates hierarchical and cross-modal discrepancies between image and class embeddings. Right: DSFR leverages the class-level feature and instance-aware attributes from class and image embeddings to refine the joint embedding.}
  % \caption{Left: HCMI mediates hierarchical and cross-modal discrepancies between image and class embeddings. Right: DSFR leverages the class-level feature from class embedding and instance-aware attributes from image embedding to refine the joint embedding.}
  \label{fig:our_module}
\end{figure*}

We use pretrained CLIP to encode the textual keypoint descriptions to get the corresponding joint embedding and our multimodal input~(textual class description and query image) to get our class embedding and image embedding.
Both class embedding and image embedding contain the query image feature while the former is more related to the general category information and the latter is more related to the specific instance feature. To this end, we proposed \textbf{Hierarchical Cross-Modal Interaction} and \textbf{Dual-Stream Feature Refinement} to enhance the joint embedding for CAPE. First, Hierarchical Cross-Modal Interaction (HCMI) amplifies dominant instance patterns in the image embedding by the class embedding and enriches class embedding with instance-adaptive details from the image embedding. Then, Dual-Stream Feature Refinement (DSFR) performs dual cross-attention between the original joint embedding and the refined class/image embedding, yielding residual-enhanced embedding. The computation of HCMI and DSFR can be described as Eq.\ref{eq:HCMI_eq} and \ref{eq:DSFR_eq}, respectively. The final enhanced joint embedding is fed into the subsequent transformer encoder-decoder module.
% \begin{equation}
%     \label{eq:HCMI_eq}
%         e^{'}_{img}, e^{'}_{cls} = HCMI(e_{img}, e_{cls}),
% \end{equation}
% \begin{equation}
%     \label{eq:DSFR_eq}
%         e^{'}_{joint} = e_{joint} + DSFR(e_{joint}, e^{'}_{img}, e^{'}_{cls}),
% \end{equation}
\begin{gather}
    \label{eq:HCMI_eq}
        e^{'}_{img}, e^{'}_{cls} = HCMI(e_{img}, e_{cls}), \\[1pt] % 手动控制两行间距
    \label{eq:DSFR_eq}
        e^{'}_{joint} = e_{joint} + DSFR(e_{joint}, e^{'}_{img}, e^{'}_{cls}),
\end{gather}
where $e_{img}$ and $e_{cls}$ represent the image embedding and class embedding encoded from CLIP, $e^{'}_{img}$ and $e^{'}_{cls}$ represent the image embedding and class embedding enhanced by HCMI. $e_{joint}$ and $e^{'}_{joint}$ represent the joint embedding encoded from CLIP and the joint embedding refined by DSFR.

Once the refined joint embedding with class and instance info have been learned, we employ a graph transformer decoder, to propagate node features through graph connections. Updated graph node features are decoded into similarity heatmaps and offset maps through parallel CNN heads. We extract the heatmap peaks and apply offset correction to obtain the final joint coordinates.

% In the following sections, we detail the Hierarchical Cross-Modal Interaction and Dual-Stream Feature Refinement, which are the core innovations that bridge our multimodal inputs with graph-based category-agnostic pose estimation. 

\subsection{Hierarchical Cross-Modal Interaction (HCMI)}
\label{sec:HCMI}
Although a certain degree of feature alignment has been done between image embedding and class embedding during the pre-training process of CLIP, there are still differences in feature representation between these two. The image embedding $e_{img}$ contains more information about the specific visual features of the query image at the lower hierarchy and introduces more image noise when used alone, while the class embedding $e_{cls}$ at the higher hierarchy contains more general and comprehensive information about the query image of that class and cannot be adapted to the specific query image more flexibly when used alone.

To bridge the hierarchy and modality gap between these two visual and semantic embeddings, we propose to interactively refine the image embedding $e_{img}$ and class embedding $e_{cls}$ through a self-attention mechanism, as illustrated in Figure \ref{fig:our_module}. This module processes two inputs: (1) one concrete query image from the visual modality and (2) universal category description from the textual modality. Both of them are independently encoded by CLIP to generate corresponding image embedding $e_{img}$ and class embedding $e_{cls}$. They are concatenated and undergo cross-modal interaction through self-attention mechanism, followed by multilayer perceptron layers~(MLP) to produce enhanced embedding. The output is spliced to produce enhanced image embedding $e^{'}_{img}$ and class embedding $e^{'}_{cls}$. This interaction is formulated as follows:
\begin{equation}
    \label{eq:HCMI}
    e^{'}_{img},e^{'}_{cls} =  \mathrm{MLP}(SelfAttn(concat(e_{img},e_{cls}))),
\end{equation}
where SelfAttn($\cdot$) denotes the self-attention operation that captures cross-modal dependencies, followed by a MLP module with ReLU-activated fully-connected layers.  $e^{'}_{img}$ and $e^{'}_{cls}$ indicate the enhanced image embedding and class embedding, which have reduced inter-modality discrepancy through the attention alignment and mutually enriched semantic-visual information via cross-modal feature propagation. Their compatibilities for subsequent joint embedding construction are also enhanced in this interaction process.

\subsection{Dual-Stream Feature Refinement (DSFR)}
\label{sec:DSFR}
Building upon the enhanced image embedding and class embedding above, the dual-stream feature refinement module dynamically adapts the CLIP-generated joint embedding to enhance cross-modal alignment between the query image's visual feature, the category feature and the support joint feature. As shown in Figure \ref{fig:our_module}, DSFR module processes three inputs: (1) the text-modality joint embedding $e_{joint} \in \mathbb{R}^{N\times C}$ derived from textual keypoint descriptions, (2) the text-modality class embedding $e^{'}_{cls} \in \mathbb{R}^{C}$ derived from image caption and refined by HCMI module, (3) the image-modality image embedding $e^{'}_{img} \in \mathbb{R}^{C}$ encoded from CLIP and refined by HCMI module. After the cross-modal feature Refinement, DSFR module outputs an augmented joint embedding $e^{'}_{joint} \in \mathbb{R}^{N\times C}$, where C denotes the embedding dimension of the LLM encoder and N represents the predefined keypoint number. 

The DSFR module employs a dual-stream cross-attention mechanism where original joint embedding serves as the query stream, while enhanced image embedding and enhanced class embedding respectively serves as the key-value stream for two parallel cross-attention computations. This is to enable original joint embedding, which is full of keypoint feature information, to dynamically retrieve instance-specific visual features from enhanced image embedding and general class-aware attributes from enhanced class embedding. This refinement can bridge textual joint priors with specific query image feature through attention-based feature correlation effectively. We use ReLU-activated MLP layers to process the outputs of the cross-attention and get the refined joint embedding. This refined joint embedding conditioned on the query image embedding and class embedding is formally denoted as: $e^{img}_{joint} \in \mathbb{R}^{N\times C}$ and $e^{cls}_{joint} \in \mathbb{R}^{N\times C}$. This attention computation can be described as follows:
\begin{equation}
    \label{eq:cross-attn-img}
    e^{img}_{joint} = \mathrm{MLP}(CrossAttn(q=e_{joint},k,v=e^{'}_{img}))
\end{equation}
\begin{equation}
    \label{eq:cross-attn-cls}
    e^{cls}_{joint} = \mathrm{MLP}(CrossAttn(q=e_{joint},k,v=e^{'}_{cls}))
\end{equation}

To incorporate the information from $e^{img}_{joint}$ and $e^{cls}_{joint}$ more flexibly, we employ two independent gating networks to adaptively weight the enhanced joint embeddings of $e^{img}_{joint}$ and $e^{cls}_{joint}$. Gating networks produce joint-wise scores of the cross-modal feature using linear layers with sigmoid activation. The refined joint embedding is subsequently computed as follow:
\begin{equation}
    \label{eq:learnableWeights}
    e^{'}_{joint} = \mathrm{MLP}(\alpha \odot e^{img}_{joint} + \beta \odot e^{cls}_{joint} + e_{joint}),
\end{equation}
where $\alpha,\beta\in[0,1]^N$ denote learnable coefficients, $\odot$ indicates Hadamard product and the MLP contains ReLU-activated hidden layers for nonlinear projection.

% 损失函数介绍
\subsection{Loss Function}
We use a heatmap loss $\mathcal{L}_{heatmap}$ and an offset loss $\mathcal{L}_{offset}$ to train our model. $\mathcal{L}_{heatmap}$ supervises the similarity maps with the shape of $H\times W$ obtained from proposal generator while $\mathcal{L}_{offset}$ supervises the decoder's localization output. They can be described as follows:
\begin{equation}
    \label{eq:heatmap-loss}
    \mathcal{L}_{heatmap} = \frac{1}{N}\sum^{N}_{i=1}\frac{1}{H\cdot W}||sigmoid(H_{i})-\hat{H}_{i}||
\end{equation}
\begin{equation}
    \label{eq:offset-loss}
    \mathcal{L}_{offset} = \frac{1}{L}\sum^{L}_{l=1}\sum^{N}_{i=1}|P^{l}_{i} - \hat{P}_{i}|
\end{equation}
where $H_{i}$ is the output similarity heatmap of the i-th joint from the proposal generator, $\hat{H}_{i}$ is the corresponding ground truth heatmap, $P^{l}_{i}$ is the output location of the l-th layer in the decoder with $L$ layers and $\hat{P}_{i}$ is the corresponding ground truth location.

The total loss for training can be formatted as follows:
\begin{equation}
    \label{eq:total-loss}
    \mathcal{L} = \lambda_{heatmap} \cdot \mathcal{L}_{heatmap} + \mathcal{L}_{offset} 
\end{equation}
where the loss weight $\lambda_{heatmap}$ for the heatmap loss is 2. 
\section{Experiment}
\begin{table*}[!t] % color
  \centering
  \begin{tabular}{@{}c|c|c|ccccc|c@{}}
    \toprule
    Methods & Img Backbone & CLIP Backbone & Split1 & Split2 & Split3 & Split4 & Split5 & Avg \\
    \midrule
    POMNet & ResNet-50 & - & 84.23 & 78.25 & 78.17 & 78.68 & 79.17 & 79.70 \\
    CapeFormer & ResNet-50 & - & 89.45 & 84.88 & 83.59 & 83.53 & 85.09 & 85.31 \\
    ESCAPE & ResNet-50 & - & 86.89 & 82.55 & 81.25 & 81.72 & 81.32 & 82.74 \\
    MetaPoint+ & ResNet-50 & - & 90.43 & 85.59 & 84.52 & 84.34 & 85.96 & 86.17 \\
    X-Pose & ResNet-50 & ViT-Base-32 & 89.07 & 85.05 & 85.26 & 85.52 & 85.79 & 86.14 \\
    SDPNet & HRNet-32 & - & 91.54 & 86.72 & 85.49 & 85.77 & 87.26 & 87.36 \\
    GraphCape & Swinv2-T & - & 91.19 & \textbf{87.81} & 85.68 & 85.87 & 85.61 & 87.23 \\
    CapeX & HRNet-w32 & ViT-Base-32 & 89.1 & 85.0 & 81.9 & 84.4 & 85.4 & 85.2 \\
    CapeX & ViT-Base-16 & ViT-Base-32 & 90.75 & 82.87 & 83.18 & 85.95 & 85.49 & 85.65 \\
    CapeX & DINOv2-ViT-S & ViT-Base-32 & 90.6 & 83.74 & 83.67 & 86.87 & 85.93 & 86.18 \\
    CapeX & Swinv2-T & ViT-Base-32 & 91.9 & 86.97 & 84.41 & 86.13 & 88.64 & 87.61 \\
    \midrule
    \textbf{CapeNext} & HRNet-w32 & ViT-Base-32 & 90.2 & 86.0 & 82.9 & 85.4 & 87.1 & 86.3 \\
    \textbf{CapeNext} & ViT-Base-16 & ViT-Base-32 & 90.84 & 86.73 & \textbf{86.5} & 82.44 & 87.91 & 86.88 \\
    \textbf{CapeNext} & DINOv2-ViT-S & ViT-Base-32 & 92.12 & 87.75 & 83.76 & \textbf{87.16} & 88.95 & 87.95 \\
    \textbf{CapeNext} & Swinv2-T & ViT-Base-32 & \textbf{92.44} & 87.31 & 85.44 & 86.47 & \textbf{90.17} & \textbf{88.37} \\
    \bottomrule
  \end{tabular}
  \caption{PCK@0.2 performance on MP-100 test dataset. CapeNext outperforms other methods in the 1-shot setting.}
  \label{tab:main-results}
\end{table*}

\begin{table}
  \centering
  \setlength {\tabcolsep}{1pt}
  \begin{tabular}{@{}c|ccccc|c@{}}
    \toprule
    Settings & Split1 & Split2 & Split3 & Split4 & Split5 & Avg \\
    \midrule
    \textbf{CapeNext} & 92.44 & \textbf{87.31} & \textbf{85.44} & 86.47 & 90.17 & \textbf{88.37} \\
    \midrule
    CapeNext w/o HCMI & 91.43 & 83.4 & 84.97 & \textbf{87.27} & 86.98 & 86.81 \\
    CapeNext w/o DSFR & 89.04 & 82.72 & 82.94 & 82.97 & 83.58 & 84.25 \\
    CapeNext w/o LW & \textbf{92.59} & 86.8 & 85.03 & 86.52 & \textbf{90.5} & 88.28 \\
    \bottomrule
  \end{tabular}
  \caption{Ablation study for our module design.}
  \label{tab:ablation-module}
\end{table}
\subsection{Implementation Details}
\subsubsection{Dataset and Metric.}
We perform experiments on MP-100 dataset \cite{xu2022pose}. The dataset consists of more than 18K images of 100 categories, including human hand, human face, human body, animal face, clothes, furniture, and vehicle. We also follow five splits for standard CAPE settings. Each split divides the dataset into non-overlap 70 training categories, 10 validation categories, and 20 testing categories. 
We also adopt the MP-100 dataset with unified skeleton definitions from GraphCape and keypoint text descriptions extended in CapeX.
We primarily report probability of correct keypoint with a threshold of 0.2 (PCK@0.2) for evaluation. We also report the PCK performance of other thresholds for comparison in our supplementary material.

\subsubsection{Training Details.} 
We use the pretrained Tiny-Swin-Transformer(Swinv2-T) \cite{liu2021swin} as our default feature extraction backbone and pretrained ViT-Base-32 \cite{dosovitskiy2020image} as the frozen CLIP backbone.  We also report results using ViT-Base-16 and HRNet-w32 as feature extraction backbones to validate the generalizability. CapeNext is trained end-to-end by Adam optimizer for 200 epochs with the batch size of 32 on one A100 GPU. The initial learning rate is $10^{-5}$, reducing by a factor of 10 at the 160th and 180th epochs, following the conventional setting.

\subsection{Main Results}
\label{sec:mainResults}
We compare CapeNext with eight SOTA methods including POMNet, CapeFormer, ESCAPE \cite{nguyen2024escape}, MetaPoint, X-Pose \cite{yang2024x}, SDPNet, GraphCape and our baseline CapeX. As Table \ref{tab:main-results} shows, with Swinv2-T as backbone, CapeNext outperforms other methods by an average of 0.76\% under the 1-shot setting, validating the efficacy of our HCMI module and DSFR module. More results other backbones are also provided. 
% It should be notice that the input image size is (224,224) due to the limit of ViT, which are smaller than other settings' (256,256). supp

% \subsubsection{Results with different PCK thresholds.}
% Following the setting of previous CAPE methods, we mainly report CapeNext's PCK@0.2 performance. Here, we also show the average PCK results of CapeNext and CapeX for the comparison with the threshold of 0.05, 0.1, 0.15, 0.2, 0.25 in Table \ref{tab:capenext-pck-th}. Our method consistently outperforms CapeX, the SOTA method, to a large margin. The performance gain is even larger at smaller threshold, exceeding it by 1.89\% in PCK@0.05.

% \begin{table*}[!t]
%   \centering
%   \begin{tabular}{@{}c|ccccc|c@{}}
%     \toprule
%     Methods & PCK@0.05 & PCK@0.1 & PCK@.15 & PCK@0.2 & PCK@0.25 & AVG \\
%     \midrule
%     CapeX & 46.62 & 71.25 & 81.96 & 87.61 & 90.95 & 75.68 \\
%     CapeNext & 48.51 & 72.66 & 83.13 & 88.37 & 91.51 & \textbf{76.83}(+1.15) \\
%     \bottomrule
%   \end{tabular}
%   \caption{PCK performance of CapeX and CapeNext with different thresholds.}
%   \label{tab:capenext-pck-th}
% \end{table*}

\subsection{Ablation Study}
\label{sec:ablationResults}
% \begin{table*}[!h]
%     \centering
%     \begin{tabular}{c|cccccc}
%         \toprule
%         Methods & keypoint typo & paw$\rightarrow$foreleg & paw$\rightarrow$foot & eye$\rightarrow$eyeball & elbow$\rightarrow$forearm & shoulder$\rightarrow$upper arm \\
%         \midrule
%         CapeX & 64.10 & 86.77 & 87.38 & 87.60 & 87.37 & 86.81 \\
%         CapeNext & \textbf{66.35}(+2.25) & \textbf{87.30}(+0.53) & \textbf{88.06}(+0.68) & \textbf{88.19}(+0.59) & \textbf{88.05}(+0.68) & \textbf{87.58}(+0.77) \\
%         \bottomrule
%     \end{tabular}
%     \caption{PCK@0.2 results on MP-100 when using different keypoint prompt templates.}
%     \label{tab:ablation-result-keypoint-noise}
% \end{table*}

\subsubsection{Effect of CapeNext Design.}
Table \ref{tab:ablation-module} represents the performance of CapeNext using HCMI, DSFR, and learnable weights (LW). 
% Hierarchical Cross-Modal Interaction module is designed to enhance the information interaction between image embedding and class embedding, thereby reducing the gap between them. This improvement facilitates more effective cross-modal refinement in the subsequent stage. 
Regarding ``CapeNext w/o HCMI", we remove the HCMI module entirely and directly feed the image embedding (obtained from CLIP image encoding) and the class embedding (obtained from CLIP text encoding) into the subsequent DSFR module. In this setting, original image embedding and original class embedding serve as the key and value, respectively, and compute cross-attention with the joint embedding as query. ``CapeNext w/o HCMI" leads to a decrease of 1.56\% in the average PCK@0.2 performance, further demonstrating its effectiveness on the reverse side.
% DSFR module leverages joint embedding to query the class embedding and image embedding for information related to the keypoints of the target subject. This process incorporates class-level semantics and instance-aware visual features that are of different modalities but are beneficial to CAPE. However, if only direct projection transformations and addition are applied, the noise present in class embedding and image embedding may even adversely affect the performance of CAPE, as shown in Table \ref{tab:ablation-module}. 
Regarding ``CapeNext w/o DSFR", we replace DSFR module with two Relu-activated linear layers that perform projection transformations. These layers project the image embedding and class embedding separately, and the results are added to the joint embedding.
For learnable weights, we replace it with a fixed value of 1, which corresponds to ``CapeNext w/o LW". LW~(learnable weights) is employed to balance the influence weights between image embeddings and class embeddings. We find that the LW lead to a modest improvement, although the effect is relatively small, with an average accuracy increase of only 0.09\%.

\begin{table}[!t]
  \centering
  \setlength {\tabcolsep}{3pt}
  \begin{tabular}{@{}c|ccccc|c@{}}
    \toprule
    Settings & Split1 & Split2 & Split3 & Split4 & Split5 & Avg \\
    \midrule
    Baseline & 91.9 & 86.97 & 84.41 & 86.13 & 88.64 & 87.61 \\
    +img emb & \textbf{92.77} & \textbf{87.36} & 84.55 & 86.31 & 89.76 & 88.15 \\
    +img \& cls emb & 92.44 & 87.31 & \textbf{85.44} & \textbf{86.47} & \textbf{90.17} & \textbf{88.37} \\
    \bottomrule
  \end{tabular}
  \caption{Ablation study for our multimodal input.}
  \label{tab:ablation-input}
\end{table}

\begin{table}[!h]
	\centering
    \setlength {\tabcolsep}{1pt}
	\begin{tabular}{c|ccccc|c}
		\toprule
		Settings &Split1 & Split2 & Split3 & Split4 & Split5 & Avg  \\
		\midrule
		Ours w/o LW & 92.23 & 86.18 & 84.64 & 84.96 & 89.91 & 87.58 \\
        Ours w/ LW & 91.94 & 86.94 & 85.48 & 86.12 & 90.27 & \textbf{88.15}(+0.67) \\
		\bottomrule
	\end{tabular}
    \caption{PCK@0.2 results when using noisy class prompts.}
    \label{tab:ablation-result-class-noise}
\end{table}

\begin{table}[!h]
    \centering
    \setlength {\tabcolsep}{1pt}
    \begin{tabular}{c|c|c|c}
        \toprule
        Methods & keypoint typo & paw$\rightarrow$foreleg & elbow$\rightarrow$forearm \\
        \midrule
        CapeX & 64.10 & 86.77 & 87.37 \\
        CapeNext & \textbf{66.35}(+2.25) & \textbf{87.30}(+0.53) & \textbf{88.05}(+0.68) \\
        \midrule
        Methods & eye$\rightarrow$eyeball & paw$\rightarrow$foot & shoulder$\rightarrow$upper arm \\
        CapeX & 87.60 & 87.38 & 86.81 \\
        CapeNext & \textbf{88.19}(+0.59) & \textbf{88.06}(+0.68) & \textbf{87.58}(+0.77) \\
        \bottomrule
    \end{tabular}
    \caption{PCK@0.2 results when using noisy joint prompts.}
    \label{tab:ablation-result-keypoint-noise}
\end{table}

\begin{figure}[!t]
  \centering
   \includegraphics[width=0.9\linewidth]{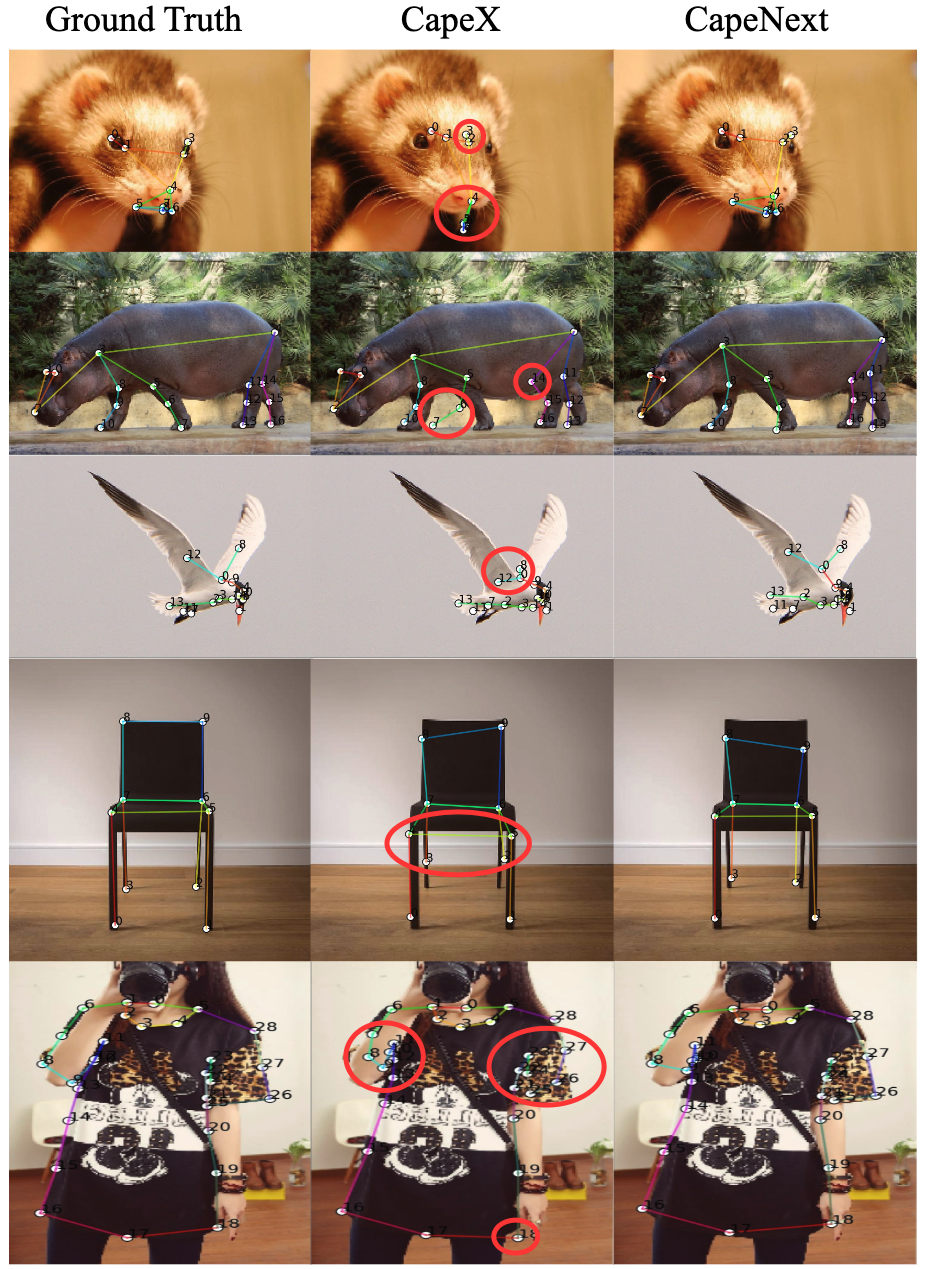}
   \caption{Qualitative results. We visualize joint predictions on MP-100 test dataset. Zoom in for more details.}
   \label{fig:qualitativeResults}
\end{figure}

\subsubsection{Insights of Multimodal Input.}
To demonstrate the effectiveness of our multimodal input data, we conducted ablation studies on both the image embedding and class embedding. The baseline setting refers to the official implementation in CapeX. For image embedding, we added its computational contributions in both HCMI and DSFR based on the baseline setting. A similar process was employed when we further incorporated the class embedding.
As shown in Table \ref{tab:ablation-input}, the introduction of image embedding significantly improved the accuracy of the model, enhancing 0.54\% average performance of PCK@0.2 over the baseline. This demonstrates that our additional image embedding has a notable positive effect on optimizing the model. Furthermore, additional class embedding further improved the model, achieving an average accuracy of 88.37\%, which further validates the effectiveness of our multimodal input.

\subsubsection{Robustness to Noisy Prompts.}
We evaluate model under two noise conditions:
(1) class prompt noise via random substitution with incorrect categories. With gating mechanism (LW), CapeNext exhibits strong class noise tolerance with merely 0.2\% PCK@0.2 drop (88.3\%→88.1\%) as shown in Table \ref{tab:ablation-result-class-noise}. We also retrain the model without gating mechanism, average PCK@0.2 performance decrease to 87.5\%, confirming the gate's error suppression capability. 
(2) keypoint prompt noise using typo patterns (e.g., ``left eey", ``nosse") from CapeX. While both CapeNext and CapeX suffer performance degradation, CapeNext mitigates feature corruption, reducing error propagation from inaccurate keypoint features. To further quantify sensitivity, we conduct multiple cross-prompt-template evaluation by modifying keypoint descriptors for evaluation. For example, from ``xx paw" to ``xx foreleg" means models are trained with ``xx paw" but tested with ``xx foreleg". PCK results in Table \ref{tab:ablation-result-keypoint-noise} show that both models suffer from this variant prompt template but CapeNext demonstrates relative robustness through cross-modal feature refinement.

\subsection{Qualitative Results}
Figure \ref{fig:qualitativeResults} presents qualitative comparisons between our method and the competing baseline CapeX. The qualitative results show that CapeNext outperforms the baseline, validating its effectiveness. Red circles highlight regions with significant differences. The examples cover animal face, animal body, furniture and clothes. However, due to the inherent complexity of the CAPE task, some joints exhibit inaccuracies (such as ``top left side of the backseat" and ``top right side of the backseat" of the chair in the forth case), highlighting the challenges in this domain.

\section{Conclusion}
In this paper, we revisit the role of support information in category-agnostic pose estimation and propose a novel framework, CapeNext, which is the first to simultaneously integrate query information and class descriptions to facilitate dynamic keypoint feature learning. We introduce two innovative modules, HCMI and DSFR, designed to enhance keypoint embeddings by fully leveraging multiple modalities. Our CapeNext not only outperforms the state-of-the-art methods by a large margin, but also maintains remarkable robustness under noisy inputs, including noisy class prompts and typo keypoint prompts.
\section{Acknowledgments}
Bo Tang and Yu Zhu were partially supported by National Science Foundation of China (NSFC No. 62422206) and a research gift from AlayaDB Inc.
Dan Zeng was supported by the National Natural Science Foundation of China (No. 62206123).
This work is supported by the National Natural Science Foundation of China (62206123, 62466013, 62176170, 62176169) and the Sichuan Science and Technology Program (2025ZNSFSC0469).

\bibliography{aaai2026}

\end{document}

\maketitle
\section{Additional Experiments}
In this section, we extend the result of the manuscript and conduct additional experiments on the proposed modules to validate our effectiveness. The experiments settings of network backbone are consistent with the settings in the manuscript. We will make our project codes public in the final version.

\subsection{Extended Main Results.}
\subsubsection{Results with Different PCK Thresholds.}
Following the setting of previous CAPE methods, we mainly report CapeNext's PCK@0.2 performance. Here, we also show the average PCK results of CapeNext and CapeX for the comparison with the threshold of 0.05, 0.1, 0.15, 0.2, 0.25 in Table \ref{tab:capenext-pck-th}. Our method consistently outperforms CapeX, the SOTA method, to a large margin. The performance gain is even larger at smaller threshold, exceeding it by 1.89\% in PCK@0.05.

\begin{table}[!h]
  \centering
  \setlength {\tabcolsep}{2pt}
  \begin{tabular}{@{}c|ccccc|c@{}}
    \toprule
    Methods & @0.05 & @0.1 & @.15 & @0.2 & @0.25 & Avg \\
    \midrule
    CapeX & 46.62 & 71.25 & 81.96 & 87.61 & 90.95 & 75.68 \\
    CapeNext & 48.51 & 72.66 & 83.13 & 88.37 & 91.51 & \textbf{76.83}(+1.15) \\
    \bottomrule
  \end{tabular}
  \caption{PCK performance of CapeX and CapeNext of different thresholds.}
  \label{tab:capenext-pck-th}
\end{table}

\subsubsection{Additional Qualitative results.}
\begin{figure*}[h]
  \centering
   \includegraphics[width=0.8\linewidth]{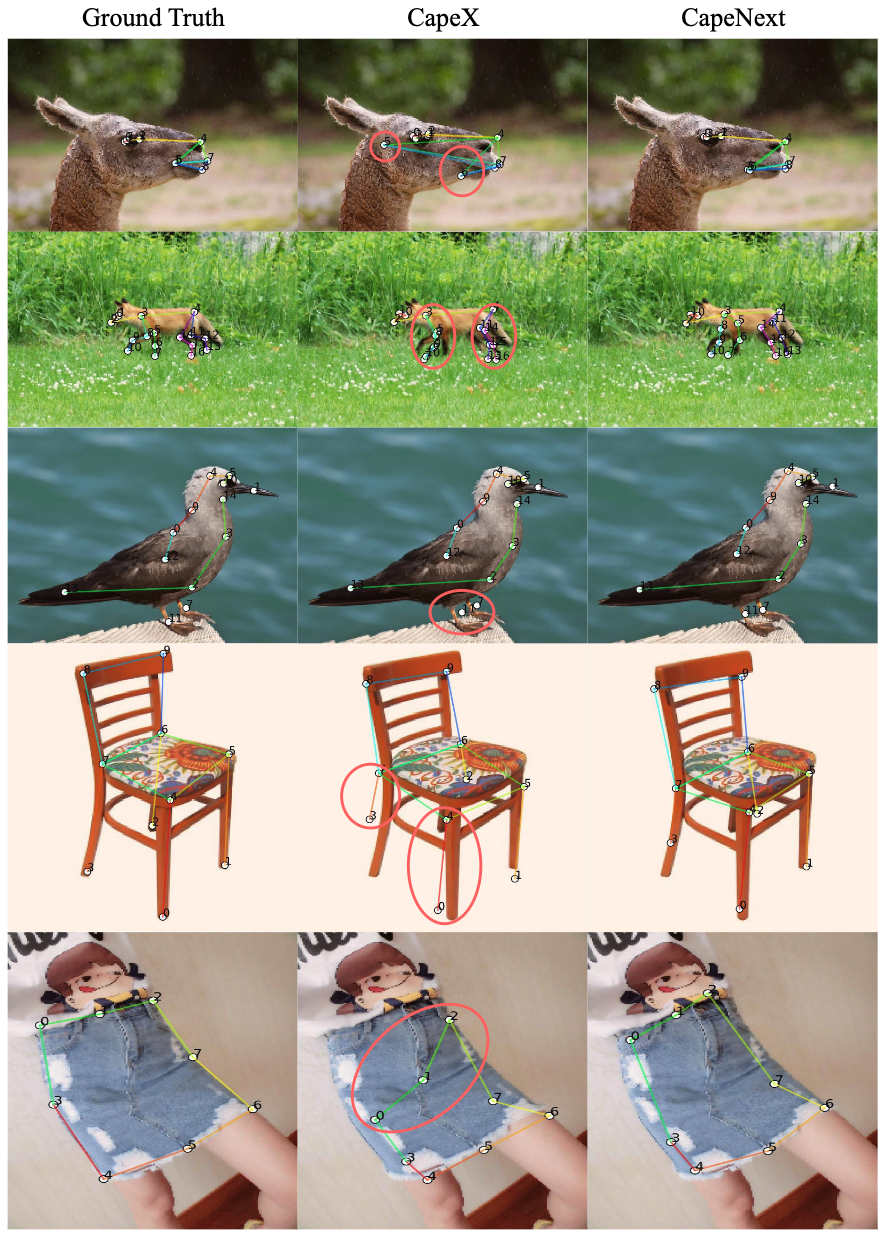}
   \caption{\textbf{Qualitative results.} We visualize additional joint predictions on MP-100 test dataset.}
   \label{fig:supple_vis1}
\end{figure*}

In order to comprehensively show the superior performance of our model, we have provided additional visualization results compared to CapeX in Figure \ref{fig:supple_vis1}, which complement and enhance the visual evidence presented earlier. 

\begin{figure*}[h]
  \centering
   \includegraphics[width=0.8\linewidth]{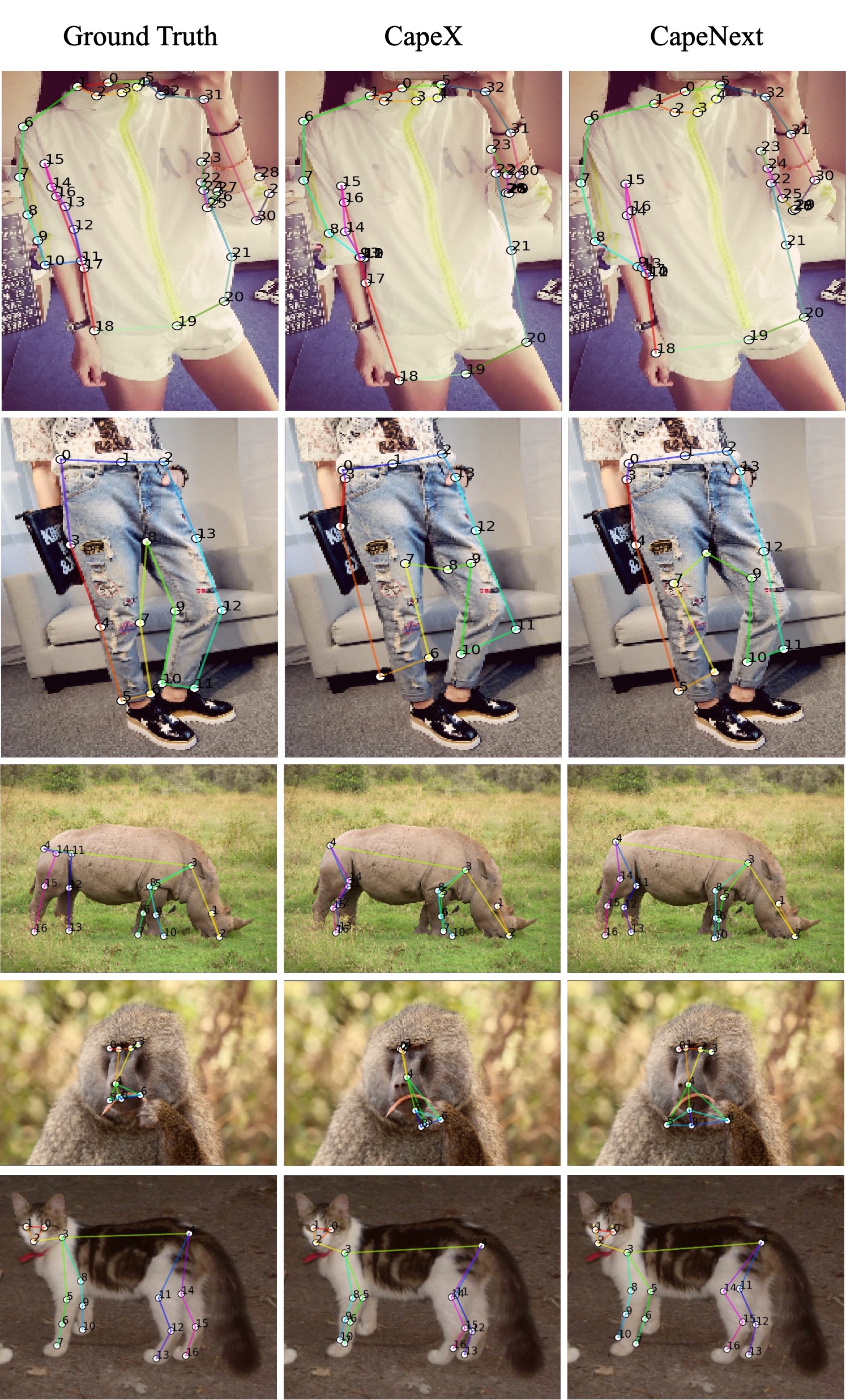}
   \caption{\textbf{Failure case.} Although the model sometimes makes inaccurate predictions, it still achieves improvements compared to the baseline.}
   \label{fig:failure_case}
\end{figure*}

Due to the challenges of the CAPE task, we also present some failure cases in Figure \ref{fig:failure_case}. Although the prediction results are inaccurate, they still show improvements compared to the baseline.

\subsubsection{Results on Sub-datasets.} % color
MP-100 is composed of 13 distinct pose estimation datasets, serving as a representative benchmark that includes 5 distinct data splits. To conduct a refined performance analysis, we first traced the original source dataset of each test sample within MP-100's 5 splits. Subsequently, we computed the performance of the models on each dataset separately for every individual split and averaged the performance scores obtained for each dataset across the 5 splits to derive the final evaluation results, which are presented in Table \ref{tab:cape_mp100_performance}.

\begin{table}[h]
\centering
\setlength {\tabcolsep}{1pt}
\begin{tabular}{l|ccccc}
\toprule
Methods   & AnimalWeb & AP-10K & CarFusion & DeepFashion & Deeppose \\
\midrule
CapeX    & 85.85     & 78.55    & 48.14     & 79.06       & 45.50    \\
CapeNext & \textbf{86.26}     & \textbf{78.92}    & \textbf{48.74}     & \textbf{79.72}       & \textbf{46.91}    \\
\bottomrule
\end{tabular}
\caption{Performance comparison between CapeX and CapeNext on constituent datasets of MP-100.}
\label{tab:cape_mp100_performance}
\end{table}

\subsubsection{Analysis of DSFR's Intermediate Embeddings and Dynamic Adjustment.}
To investigate DSFR’s dynamic adjustment capability for image and text embeddings, we conducted experiments on the MP-100 test set. Taking Split-1 as an example, we randomly sampled 300 cases from this split and computed the average cosine similarity between DSFR’s inputs (joint embedding, image embedding, text embedding) and its output enhanced joint embedding. As shown in Table \ref{tab:dsfr_cosine}, when incorrect category descriptions are used as text inputs, the cosine similarity of text embeddings decreases (4.22\% vs. 3.95\%), while the image embeddings increases (1.11\% vs. 1.96\%). This observation directly confirms DSFR’s ability to dynamically adjust the contribution of image and text embeddings based on input quality.
\begin{table}[!h] % color
  \centering
  \begin{tabular}{l|c|ccc} 
    \toprule
    Settings & Noise Type & joint & text & img \\
    \midrule
    CapeNext & - & 81.14\% & 4.22\% & 1.11\%  \\
    CapeNext & random cls & 81.09\% & 3.95\%$\downarrow$ & 1.96\%$\uparrow$ \\
    \bottomrule
  \end{tabular}
  \caption{Cosine similarity score between DSFR's inputs and output when using different category description text.}
  \label{tab:dsfr_cosine}
\end{table}

\begin{table*}[!h]
	\centering
	\begin{tabular}{c|c|ccccc|c}
		\toprule
		Methods & Species & PCK@0.05 & PCK@0.1 & PCK@0.15 & PCK@0.2 & PCK@0.25 & Avg  \\
		\midrule
		CapeX & animal body & 44.28 & 73.84 & 86.39 & 92.01 & 94.84 & 78.27 \\
        CapeNext & animal body & 46.12 & 74.52 & 86.80 & 92.19 & 94.94 & \textbf{78.91}(+0.64) \\
        \midrule
        CapeX & cat body & 44.61 & 71.85 & 85.58 & 91.91 & 94.89 & 77.77 \\
        CapeNext & cat body & 45.74 & 74.37 & 87.10 & 92.42 & 95.33 & \textbf{78.99}(+1.22) \\
		\bottomrule
	\end{tabular}
    \caption{PCK performance of different thresholds on specific super-category and category from MP-100 test dataset.}
    \label{tab:ablation-result-polysemy-category}
\end{table*}

\begin{table*}[!h] 
	\centering
	\begin{tabular}{c|c|ccccc|c}
		\toprule
		Methods & Noise Type & Split1 & Split2 & Split3 & Split4 & Split5 & Avg  \\
		\midrule
        CapeX & keypoint typo & 64.01 & 65.32 & 59.51 & 64.24 & 67.41 & 64.10 \\
        CapeNext & keypoint typo & 64.39 & 68.37 & 64.87 & 65.84 & 68.26 & \textbf{66.35}(+2.25) \\
        \midrule
        CapeX & paw\textrightarrow foreleg & 90.83 & 86.6 & 83.75 & 85.1 & 87.57 & 86.77 \\
        CapeNext & paw\textrightarrow foreleg & 90.21 & 86.6 & 85.07 & 85 & 89.61 & \textbf{87.30}(+0.53) \\
        \midrule
        CapeX & paw\textrightarrow foot & 91.55 & 86.91 & 84.34 & 85.86 & 88.24 & 87.38 \\
        CapeNext & paw\textrightarrow foot & 91.82 & 86.93 & 85.43 & 86.11 & 90 & \textbf{88.06}(+0.68) \\
        \midrule
        CapeX & eye\textrightarrow eyeball & 91.9 & 86.97 & 84.39 & 86.11 & 88.62 & 87.60 \\
        CapeNext & eye\textrightarrow eyeball & 92.01 & 86.95 & 85.5 & 86.29 & 90.18 & \textbf{88.19}(+0.59) \\
        \midrule
        CapeX & elbow\textrightarrow forearm & 91.69 & 86.89 & 84.11 & 85.83 & 88.32 & 87.37 \\
        CapeNext & elbow\textrightarrow forearm & 92.11 & 86.82 & 85.27 & 86.02 & 90.05 & \textbf{88.05}(+0.68) \\
        \midrule
        CapeX & shoulder\textrightarrow upper arm & 91.53 & 86.62 & 82.99 & 85.25 & 87.68 & 86.81 \\
        CapeNext & shoulder\textrightarrow upper arm & 92.04 & 86.7 & 84.04 & 85.65 & 89.45 & \textbf{87.58}(+0.77) \\
		\bottomrule
	\end{tabular}
    \caption{PCK@0.2 results on MP-100 when using different keypoint prompt templates.}
    \label{tab:ablation-result-keypoint-noise}
\end{table*}

\begin{table*}[!h] % color
  \centering
  \begin{tabular}{l|c|ccccc|c} 
    \toprule
    Methods   & Noise Type       & Split1 & Split2 & Split3 & Split4 & Split5 & Avg    \\
    \midrule
    CapeNext & -                & 92.44  & 87.31  & 85.44  & 86.47  & 90.17  & \textbf{88.37}  \\
    CapeNext & null cls string  & 91.85  & 86.95  & 85.50  & 85.98  & 90.30  & 88.116 \\
    CapeNext & random cls string & 91.95  & 86.95  & 85.48  & 86.13  & 90.27  & 88.156 \\
    \bottomrule
  \end{tabular}
  \caption{PCK@0.2 results on MP-100 when using different category description text.}
  \label{tab:capenext_noise_performance}
\end{table*}

\begin{figure}[!h]
  \centering
   \includegraphics[width=0.98\linewidth]{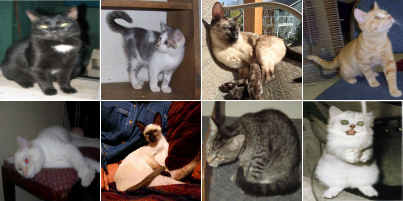}
   \caption{Cats in MP-100 with large intra-class variations.}
   \label{fig:ablation-result-polysemy-category}
\end{figure}

\subsection{Polysemy Handling and Fine-grained Category Control Evaluation.}
We check the performance of our method and CapeX when meeting the polysemy issue. We evaluate models on the super-category of animal body(average PCK results on 5 splits of different thresholds) in Table \ref{tab:ablation-result-polysemy-category}. The categories under ``animal body" share identical keypoint definitions and prompts but exhibit distinct visual cues due to inherent category differences (e.g., dog, horse, rabbit). This makes them well-suited for verifying the model's ability to resolve polysemy issue when keypoint texts are identical across different categories.

To examine the model's capability for fine-grained category control, we evaluate the performance of our method and CapeX on the "cat body" category. Below cat examples (standing black cat vs. curled white cat) show the large intra-class variations in Figure \ref{fig:ablation-result-polysemy-category}. Added experiments show that ours solves these issues better.

\subsection{Detailed Results of Robustness to Noisy Keypoint Templates and Category Description Text.}
To verify the model's adaptability and robustness when using incorrect keypoint prompts or different keypoint prompt templates, we conducted experiments on MP-100, comparing the performance of CapeX and CapeNext (same as the manuscript). Extended results are presented in Table \ref{tab:ablation-result-keypoint-noise}. It is evident that CapeNext outperforms CapeX both when using misspelled keypoint prompts and when employing keypoint prompt templates different from those used during training.

% color
We also replaced the category information in the category description text with null values and random categories during testing. The results in Table \ref{tab:capenext_noise_performance} show that the model's PCK metric decreases minimally, demonstrating the model's robustness against erroneous category description text.

\subsection{Effect of Refinement Module.}
\subsubsection{Layer Number Experiment.} % color
Our proposed innovative module of HCMI and DSFR can be encapsulated into an independent refinement layer. To explore the optimal usage of this layer, we conducted experiments to investigate the effects of stacking it multiple times. This layer takes three inputs: image embedding $e_{img}$, class embedding $e_{cls}$, and joint embedding $e_{joint}$, and produces an enhanced joint embedding $e^{'}_{joint}$. When stacking multiple layers, each subsequent layer takes the original inputs $e_{img}$ and $e_{cls}$, along with the enhanced output $e^{'}_{joint}$ from the previous layer, and generates a further refined output $e^{''}_{joint}$. Formally, the operation of the i-th layer can be described as follows:
\begin{equation}
    \label{eq:ablation-layerNum}
    e^{i}_{joint} = Layer(e_{img},e_{cls},e^{i-1}_{joint}),
\end{equation}
where $e^{0}_{joint}$ = $e_{joint}$ represents the initial input, and $e^{i}_{joint}$ denotes the enhanced output after the i-th layer.

\begin{table}[!h]
  \centering
  \setlength {\tabcolsep}{2pt}
  \begin{tabular}{@{}c|ccccc|c@{}}
    \toprule
    Layer Number & Split1 & Split2 & Split3 & Split4 & Split5 & Avg \\
    \midrule
    0 & 91.9 & 86.97 & 84.41 & 86.13 & 88.64 & 87.61 \\
    \textbf{1} & 92.44 & 87.31 & 85.44 & 86.47 & 90.17 & \textbf{88.37} \\
    2 & 92.42 & 86.38 & 85.8 & 86.6 & 89.95 & 88.23 \\
    3 & 92.62 & 85.8 & 85.44 & 86.57 & 89.53 & 87.99 \\
    \bottomrule
  \end{tabular}
  \caption{Ablation study for the layer number. ``0" corresponds to the baseline and ``1" corresponds to CapeNext setting. ``2",``3" corresponds to repeating the joint embedding refinement operation two to three times, respectively.}
  \label{tab:ablation-layerNum}
\end{table}

\begin{table}[!h]
  \centering
  \begin{tabular}{l|cc} 
    \toprule
    Methods  & Params(M) & Avg   \\
    \midrule
    CapeX & 100.7M           & 87.61 \\
    CapeX+Trans.Encoder & 120.7M & 86.96 \\
    CapeNext & 116.5M        & \textbf{88.37} \\
    \bottomrule
  \end{tabular}
  \caption{PCK@0.2 performance of different methods.}
  \label{tab:model_avg_performance}
\end{table}

The corresponding experimental results in Table \ref{tab:ablation-layerNum} demonstrate that the model achieves its best performance when this layer is stacked only once. Specifically, with a single layer, the model achieves a PCK@0.2 accuracy of 88.37\% when the model achieves a lower average accuracy with double or triple layers. This is because a single enhancement layer strikes the right balance between enriching the joint embedding with additional class/instance information and preserving the integrity of the original keypoint features. Further enhancements risk introducing redundancy or suppressing critical feature expressions, which could degrade the model performance. Therefore, according to Equation \ref{tab:ablation-layerNum}, enhancing joint embedding with multimodal embedding only once proves to be the most appropriate for our model in the MP-100 benchmark.

\subsubsection{Module Experiment.} % color
We replaced the HCMI/DSFR module with a standard Transformer Encoder to identify the source of performance gains. Results in Table \ref{tab:model_avg_performance} confirm that our proposed new module is the core driver behind the improved performance.

\subsection{Ablation Study on Image Masking.}
% 待更名
To investigate the potential of image embedding in improving CAPE, we also considered applying a mask to the image using the bounding box information provided in the dataset. 

Specifically, we set the pixels outside the bounding box to zero, retaining only the target instance. This process aims to minimize noise unrelated to the target instance in the images, thereby increasing the proportion of effective information related to the target instance in the image embedding. However, it is worth-noting that for face-related data, the mask also obscures significant portions of the body that are relevant to the subject, leaving only the facial region.

We conducted an ablation study for our multimodal input data in this setting. The experimental settings, including the backbone network, remains consistent with previous settings, except for the mask process during training and test process. As we can see in Table \ref{tab:maskResults}, the application of mask processing improves the model's accuracy across multiple settings. Specifically, the model employing our multimodal input achieves the highest accuracy, reaching 88.53\% in terms of PCK@0.2. However, it should be note that since mask processing was applied during both training and testing, these results are not directly comparable to the MP-100 benchmark.

\begin{table}[h]
  \centering
  \setlength {\tabcolsep}{1pt}
  \begin{tabular}{@{}c|ccccc|c@{}}
    \toprule
    Settings & Split1 & Split2 & Split3 & Split4 & Split5 & Avg \\
    \midrule
    baseline & 92.59 & 87.53 & 83.85 & 87.2 & 88.46 & 87.92 \\
    baseline + $e_{cls}$ & 92.51 & 87.17 & 84.28 & 86.47 & 89.84 & 88.05 \\
    baseline + $e_{img}$& 92.56 & 87.4 & 83.63 & 86.54 & 88.63 & 87.75 \\
    \textbf{CapeNext} & 92.24 & 87.58 & 85.57 & 87.15 & 90.13 & \textbf{88.53} \\
    \bottomrule
  \end{tabular}
  \caption{Quantitative results using mask operation. We show the performance of the models when different inputs are used and the query images are masked.}
  \label{tab:maskResults}
\end{table}

\begin{table}[!h]
	\centering
    \setlength{\tabcolsep}{1pt}
	\begin{tabular}{c|ccccc|c}
		\toprule
		Settings & Split1 & Split2 & Split3 & Split4 & Split5 & Avg  \\
		\midrule
		Ours w/o SuppImg & 92.44 & 87.31 & 85.44 & 86.47 & 90.17 & 88.37 \\
        Ours w/ SuppImg & 92.31 & 87.48 & 85.59 & 86.29 & 90.3 & \textbf{88.39} \\
		\bottomrule
	\end{tabular}
    \caption{PCK@0.2 results on MP-100 when using additional support image.}
    \label{tab:ablation-result-support-image}
\end{table}

\subsection{CapeNext With Support Image.}
To explore whether including a support image can lead to furthe improvements, we encode one support image and the query image with CLIP to get their embeddings and fused them with gating mechanism mentioned in DSFR module as the new query image embedding. This new query image embedding will be taken as one input of DSFR module to perform the subsequent computation, same as the original CapeNext. PCK results are shown in Table \ref{tab:ablation-result-support-image}, indicating that our multi-modal inputs are almost enough for CAPE task, while the support images are expensive to collect.

\subsection{Error-induced Scenario Predictions.}
\begin{figure*}[t]
  \centering
   \includegraphics[width=0.8\linewidth]{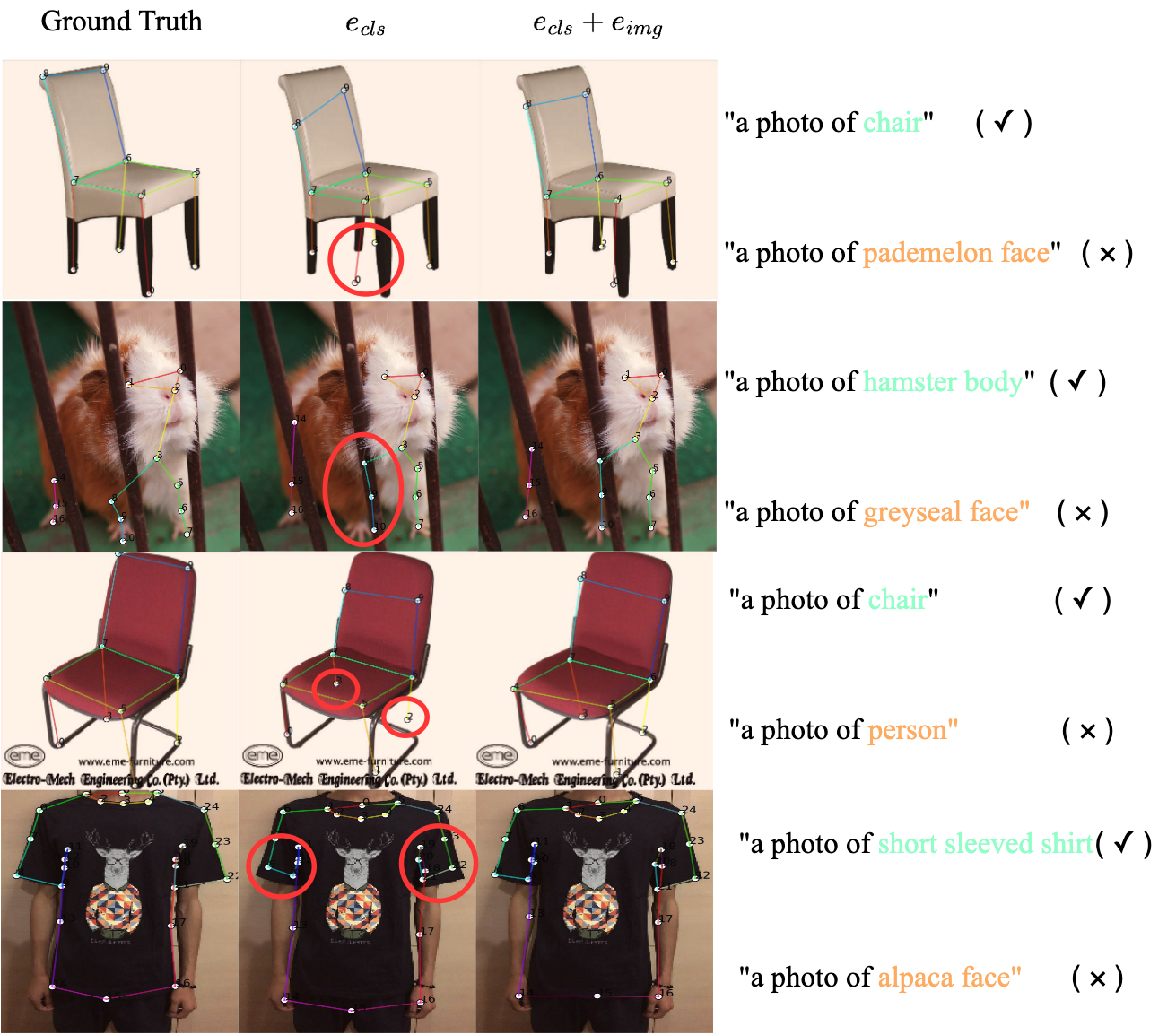}
   \caption{\textbf{Qualitative results.} We visualize joint predictions when using incorrect class descriptions on MP-100 test dataset. The second column shows the prediction with only incorrect class descriptions as input, followed by the predictions with multimodal input. Zoom in for more details.}
   \label{fig:qualitativeResultsCls}
\end{figure*}

To rigorously validate the effectiveness of image embeddings and model robustness, we also show the qualitative results comparison between ours with $e_{img}$ and $e_{cls}$ and ours but only with $e_{cls}$ in Figure \ref{fig:qualitativeResultsCls}. The textual class information is randomly replaced with the wrong class. For example, the text prompt ``a photo of chair" for a photo of a chair may be replaced with ``a photo of pademelon face" where the incorrect class (``pademelon face") is randomly sampled from MP-100 to create intentional semantic mismatching. Visualization results in the second column show that when exclusively employing incorrect class embedding $e_{cls}$, the predicted keypoints exhibited substantial positional deviation from the ground truth, particularly out of target body regions. The introduction of $e_{img}$ effectively mitigates this problem, offsetting the semantic ambiguity in the incorrect class embedding, demonstrating the robustness of CapeNext in this error-induced scenario, as shown in the third column.

\begin{table}[!h]
\centering
\setlength{\tabcolsep}{2pt}
\begin{tabular}{l|c|ccl}
\toprule
Methods   & Backbone    & Params (M) & Time (s) & Avg        \\
\midrule
CapeX    & Swinv2-T    & 28.35      & 603      & 87.61       \\
\midrule
CapeNext & HRNet-w32   & 28.54      & 774      & 86.30       \\
CapeNext & Vit-Base-16 & 86.39      & 719      & 86.88       \\
CapeNext & DINOv2-ViT-S    & 22.06      & 544      & 87.95       \\
CapeNext & Swinv2-T    & 28.35      & 676      & \textbf{88.37}       \\
\bottomrule
\end{tabular}
\caption{PCK@0.2 performance and overhead of different backbones.}
\label{tab:backbone_overhead}
\end{table}

\subsection{Computational Overhead.} %color
To assess CapeNext’s computational overhead, we compared it with baseline CapeX via one-epoch training on an NVIDIA TITAN Xp (batch size=8). With Swinv2-T as image backbone, CapeNext (116.5M params) requires 10\% more GPU memory (3588MB) and longer computation time (676.7s) than CapeX (100.7M params, 3293MB, 606.3s), but delivers superior performance (verified in subsequent sections). 

We also tested the computational overhead of other backbones in Table \ref{tab:backbone_overhead}. Results show lightweight backbones (DINOv2-ViT-S: 22.06M params, 544s) still achieve high Avg PCK@0.2 (87.95\%). Swinv2-T (28.35M params, 676s) hits the highest Avg PCK@0.2 (88.37\%), while heavier Vit-Base-16 (86.39M params) gives no proportional gains, validating our backbone selection and CapeNext’s efficiency.

% \bibliography{aaai2026_supp}